\documentclass[review]{elsarticle}

\usepackage{lineno,hyperref}
\modulolinenumbers[5]
\usepackage{amsthm}
\usepackage{amsmath}

\usepackage{subfig}
\usepackage{xcolor}

\journal{}

\begin{document}

\begin{frontmatter}

\title{Object Localization Through a Single Multiple-Model Convolutional Neural Network with a Specific Training Approach}

\author{Faraz Lotfi\tnoteref{Corresponding author}, Farnoosh Faraji, Hamid D. Taghirad}
\address{Advanced Robotics and Automated Systems (ARAS), Industrial Control Center of Excellence, Faculty of Electrical Engineering, K. N. Toosi University of Technology, Tehran 1969764499, Iran}
\tnotetext[Corresponding author]{Corresponding author.\\E-mail addresses: F.lotfi@email.kntu.ac.ir (F. Lotfi), Taghirad@kntu.ac.ir (H.D. Taghirad).}




\begin{abstract}
Object localization has a vital role in any object detector, and
therefore, has been the focus of attention by many researchers. In
this article, a special training approach is proposed for a light
convolutional neural network (CNN) to determine the region of
interest (ROI) in an image while effectively reducing the number of
probable anchor boxes. Almost all CNN based detectors utilize a
fixed input size image, which may yield poor performance when
dealing with various object sizes. In this paper, a different CNN
structure is proposed taking three different input sizes, to enhance
the performance. In order to demonstrate the effectiveness of the
proposed method, two common data set are used for training while
tracking by localization application is considered to demonstrate its
final performance. The promising results indicate the applicability
of the presented structure and the training method in practice.
\end{abstract}


\end{frontmatter}


\section{Introduction}
\textcolor{black}{Nowadays, several sensors are employed to make a system capable of realizing and handling a critical situation. Computational load and expenses, on the other hand, impose employment of the optimum type and amount of sensors. In an automated robotic system, most of the time a monocular camera may be utilized to provide rich information about the obstacles or intended targets around at a low cost. As a result, object detection through image processing has found its permanent attention among the researches.} By introducing new hardware capable of executing deep CNNs, deep learning became the key component of
almost all intelligent methods in this field. Unsurprisingly, there
is plenty of nominated research works in the field of object
detection which is entirely based on CNNs. These approaches may be
divided into single and two-stage categories. Single-stage detectors
are faster due to having just a single task to localize and
recognize objects in an image. Two-stage detectors, however, are reported to be more
accurate but not as fast as single-stage ones. Generally, the idea
behind these detectors is to perform the detection through two
stages. Firstly, an approach is performed to determine and identify the region
where the object may highly exist. Then, a
classifier window is slid on that region, considering several anchor boxes. \textcolor{black}{A major issue behind these two detectors is the trade-off between speed and accuracy. In this article, this challenging problem has been addressed by suggesting applicable approaches in designing a fast and precise object localizer. Furthermore, to demonstrate its capabilities, a tracking application is considered in which both of the mentioned characteristics become critical in performing perfect object tracking.
In what follows, some nominated researches are presented regarding the two mentioned approaches. Then, the key concepts of the proposed object localizer are elaborated. }
\subsection{Related Works}
In the literature, several structures are presented
within the single-stage detecting framework such as single-shot
multi-box detector (SSD)~\cite{SSD}, RetinaNet~\cite{RetinaNET}, and
you only look once (YOLO)~\cite{YoloV3}. The latter mentioned method
divides the incoming image into grid cells and uses a specific lost
function to perform object detection by processing the image only
once. Regarding single-stage detectors,~\cite{LeNet} is the earliest
convolutional neural networks developed when the hardware was not as
powerful as today. This CNN is using three convolutional layers, and
one of its main features is, not to connect all the outputs of the
second pooling layer to the third convolutional layer. By this
means, while the computational cost is reduced, the convolutional
filters could learn different features from the data structure.

Another approach has been presented in~\cite{AlexNet} in which the
implementation of $relu$ functions instead of $sigmoid$ has been
proposed as an innovation. This will accelerate the training process
and enhance the trainability of the neural network. Furthermore, to
improve the results, $overlapping-pooling$ has been used instead of
the common $pooling$ layers. VGG has been proposed as another
approach~\cite{VGG}, in which by increasing both the layers and the
trainable parameters, better outcomes have been reported compared to
that of the previous structures. Finally,
references~\cite{InceptionV1} and~\cite{InceptionV3} mainly focus on
employing different filter sizes such as $(1\times 1)$ and $(5\times
5)$ or even $(7\times 7)$. Moreover, in~\cite{InceptionV3}
non-square filters have been used.

Since objects may appear with various sizes and aspect ratios
two-stage detectors will completely scan the image with multi-scale
sliding windows. Regarding this concept, region interest CNN (R-CNN) is
suggested in \cite{RCNN}, resulting in more than $30\%$ improvement
in the mean average precision (mAP) of the results. In R-CNN
by using selective search approach \cite{selective_search}, $2k$
proposal regions are applied to accurately and quickly search an
image, which reduces the search space. Furthermore, feature vectors
will be extracted for each region by a CNN, and positive and
negative scores (for objects and background) will be set with a
pre-trained linear SVM. By this means the required bounding box is
determined for classification.

R-CNN resizes the images into proposal regions with a fixed size
since fully-connected (FC) layers require fixed-size images for
processing. This may yield unpleasant geometric distortions. To
remedy this problem, SPP-Nets are developed with a different
architecture to handle the issue~\cite{SPP_Net}. Needing large
storage while processing the whole image with SPP-Net, fast R-CNN
produces a feature map that helps to gather one fixed-length feature
vector for each region. As it is remarked, the feature vectors are
applied to estimate both the bounding box locations and the related
class \cite{Fast_RCNN} (it includes $C$ classes for the objects and
one class for the background). Regarding the region proposal
computation, region proposal network (RPN) is introduced in faster
R-CNN approach~\cite{Faster_RCNN}. In this network, the
convolutional features will be shared with the detector to
significantly reduce the training time. Despite the accuracy and
high speed of the faster R-CNN, it faces various problems while
dealing with small scale objects in the images. For instance, it has
major limitations over the COCO data set~\cite{COCO}, which contains
a wide range of objects on different scales. To handle this major
problem combination of complementary information of multiple sources
has been suggested in the literature~\cite{Multitask_learning1,
Multitask_learning2, Multitask_representation}.
\subsection{\textcolor{black}{The Key Concepts}}
The bottleneck of the above-mentioned approaches is the number of the necessary
anchor boxes in terms of the computational cost. Thus, one way to
enhance the performance of such detectors is to identify the ROI as
accurately as possible and to determine the center point of the
object bounding box precisely. Furthermore, needing a specific
output size, almost all the detectors use a fixed size image as the
input. This will reduce the performance and demands a deeper
structure for more various sizes. As a representative, even for
YOLOv3 as a strong approach applicable to a wide range of
applications, it has been suggested to change the structure and
anchor boxes to detect small objects or specific ones more
accurately. Considering this drawback, it is essential to have a
flexible structure for a general-purpose network, to detect a wide
range of objects in different sizes. Of course, it is challenging to
properly train such a neural network; thus, \textcolor{black}{a complete framework may be very beneficial to be presented to
further make it easier for the researchers to design and test their
own structures.}

The main contribution of this paper is firstly, to propose a new
approach in training deep CNNs to precisely localize objects in an
image. Secondly, a single multiple-model is presented to handle the
object localization for a wide range of object sizes without the need
for changing the neural network input size. The code has been
implemented such that, it would be easy to develop further
structures. Moreover, in this paper, all the underlying concepts
mentioned in the previous researches are used to increase the
performance of the proposed network. The outcome of this research
can be used either for a two-stage detector in multi-object
detection or a single object detector individually. Moreover, to
demonstrate the effectiveness of the suggested approach, a tracking
application has been considered where the proposed tracker is
compared to the eight commonly used OpenCV trackers~\cite{OpenCV}.
As a result, the performance of the suggested tracker is closely
comparable to the others and better in some situations.

The rest of this article is organized as follows. In the second
section, the structure will be presented in detail, and the
challenges are remarked together with their suggested solutions. In
section three, the method for the training process will be
elaborated as well as the issue of producing the desired data set.
Section four focuses on the results and performance analysis.
Finally, the concluding remarks are given in the last section.

\section{The Proposed Structure}
This section describes the proposed architecture of the network used
for diverse input sizes. As it has been mentioned before, this
structure is different from the commonly presented ones, capable of
handling three various input sizes. When an image enters the network
as the input, according to its size it will be processed through the
specific convolutional layers to produce the output. Unlike common
CNN architectures, the image will not penetrate all the layers
of the network, and this structure can handle multiple models inside
a single convolutional model. By this means, there is no need to
change the structure of the network to deal with small or large
objects. Two important notes are to be considered here. The output
size has to be fixed to avoid problems in the training process,
therefore, we considered three different image sizes in this paper.
Furthermore, it is beneficial to have common layers between the
three models. This will reduce the number of required trained
variables, and lead to an optimized structure. This method is much
more beneficial than using multiple distinct models.
\begin{figure}[]
   \centering
   \includegraphics[width=3.5in]{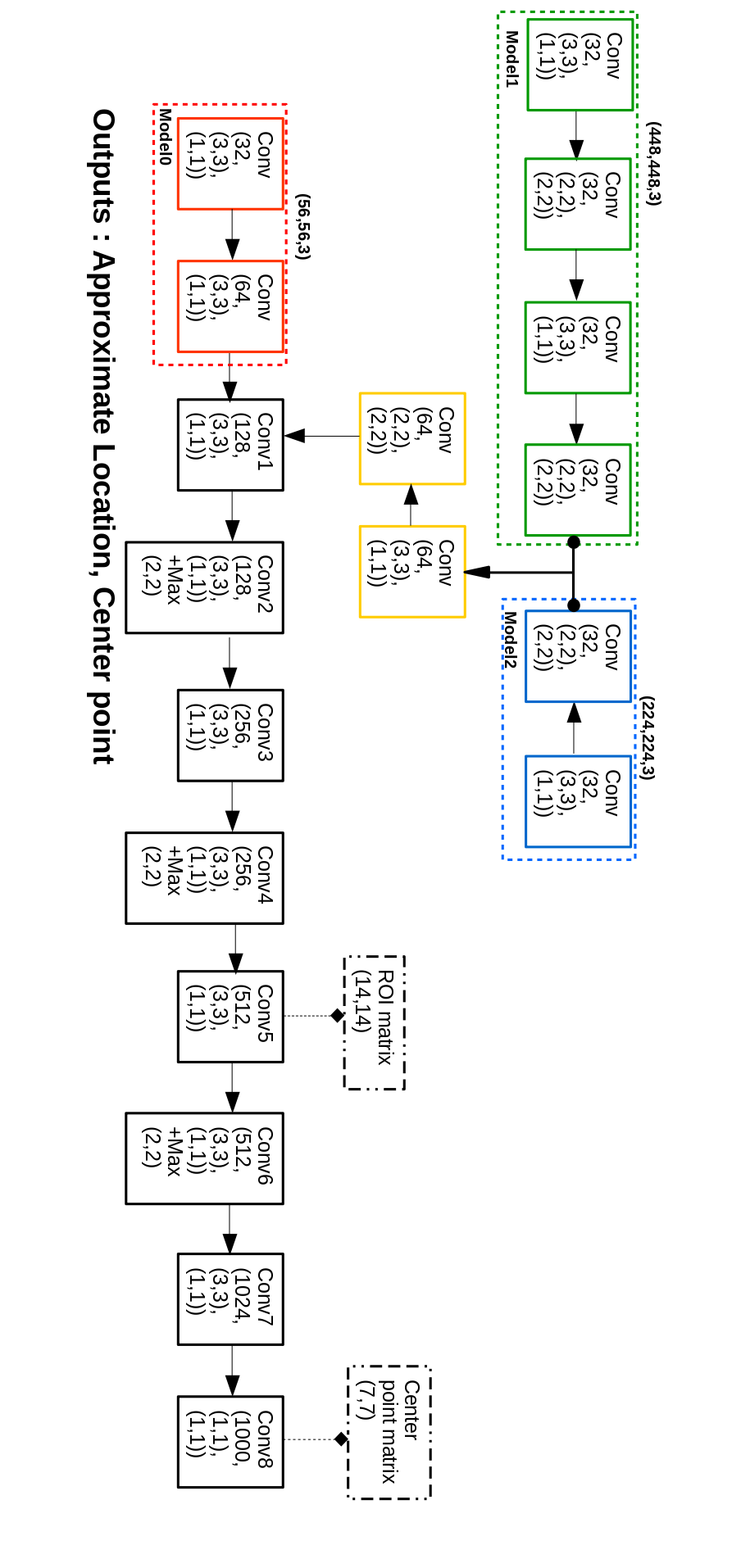}
   \caption{\small{The proposed flexible structure.}}
   \label{fig1}
\end{figure}

Regarding the aforementioned points, a flexible structure consisting
of both distinct and common convolutional layers is suggested in
this paper as illustrated in Fig.~\ref{fig1}. As it can be seen in
this figure, three inputs of $(448,448)$, $(224,224)$, and $(56,56)$
are considered for the network. For each input image, considering the
closest size, the corresponding distinct convolutional layers will
be processed through. The main underlying idea here is to downsample
the image while not losing indispensable information. In this
network for an input image with size over $(336,336)$, the green
blocks branch of the CNN will be activated, while both the blue and
the red branches became inactive. Furthermore, the yellow and the
black branches perform the processing to result in the desired
output. Note that, the black branch is common, and in terms of
using convolutional layers, the longest among all the layers. Since
having long enough common layers is necessary to reach the
required performance. There are two outputs for the network, namely
the ROI and center point matrices to be used in training this structure whose details are presented in the next section. Note that the "Max" notation used in the blocks represent the max-pooling layers following the convolutional ones.

Remark 1: There are convolutional layers with $(2,2)$ sizes and
$(2,2)$ strides. These are responsible to perform something more
than just max-pooling. However, max-pooling layers may be used,
while slightly reducing the performance.

Remark 2: In the common layers, max-pooling layers are used rather
than convolutional ones with $(2,2)$ strides. This is mainly
considered to reduce both the number of trainable variables and
structural complexity.

The presented structure represents the main idea behind having
multiple models in a single convolutional model. Depending on the
application, one may extend this structure to have more distinct
branches performing specific processing on inputs with different
sizes. Hence, the proposed structure can be employed for other image
processing purposes, by adjusting the loss function used for the
optimization.

\section{The Training Approach}
In this section, an effective approach is presented to train a
neural network for object localization. The main purpose of the
proposed convolutional neural network is to precisely identify an
ROI and the center point of the object. Considering the underlying
concepts used for effective training, some methods have been tested
and evaluated to enable such a light CNN  in terms of accuracy and speed, perform well in practice.

\subsection{The Concept}
Taking the structure presented in section 2, there are some
challenges in training the distinct layers. In fact, except for the common black branch which is always active for all the input sizes,
the gradient has to be backpropagated only to the corresponding
activated layers.
It is undeniable that those inactive layers will have a zero
gradient for the irrelevant input. When it comes to the
implementation, however, it would be challenging to count no
gradient support for those layers. This issue has been solved by
considering distinct trainable variables in the training process.
For each input image, just the trainable variables of the
corresponding activated layers and the common layers are updated,
and the other passive layers will remain unchanged. By this means, a
training process could be implemented for such flexible structures.
In the implemented code the training is performed differently for
various image sizes.

\begin{figure}[]
    \centering
    \includegraphics[width=3.5in]{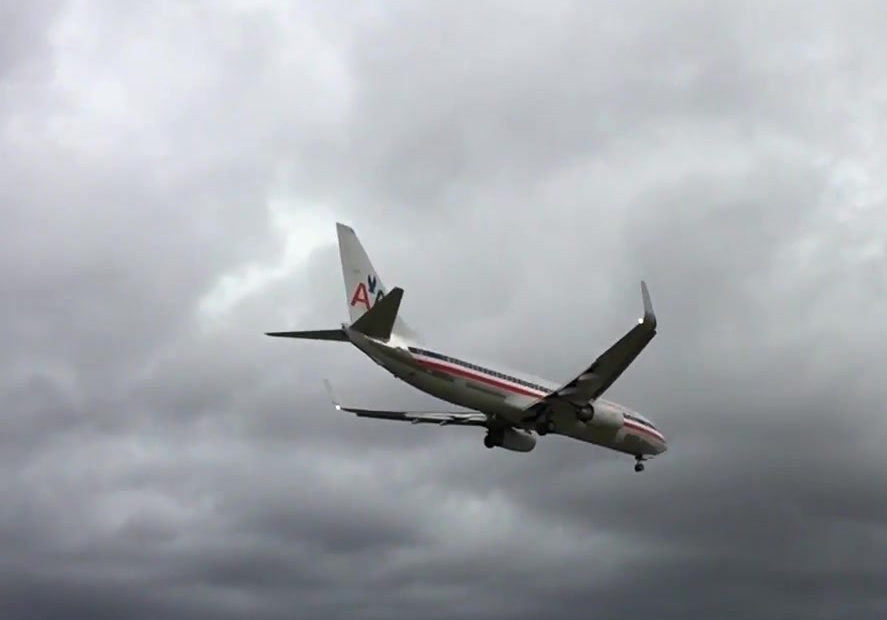}
    \caption{\small{A sample image for describing the training concept.}}
    \label{fig2}
\end{figure}
As Fig.~\ref{fig1} shows, the arrangement of the features in the
feature maps are not changed through the layers. Therefore, an
object in a specific location will trigger the same place of the
feature maps as the image is processed. By using this key concept,
it is possible to train the CNN by representing the object location
with a location matrix. This is similar to considering a mask for
each image. However, these masks are different from those
used in segmentation applications like~\cite{MRCNN}. To clarify,
here there is no need to classify each pixel in the image, and it is
sufficient to assign the pixels belonging to the corresponding
object. This is done by taking a matrix with binary values, whose
one elements correspond to the place where the object exists. In the
following example, this idea is visualized. Consider Fig.~\ref{fig2}
as the incoming image, and divide it into a $(14 \times 14)$ grid.
Take a $(14 \times 14)$ matrix, whose component is equal to one if a
part of the object is located at that position, and zero if not. As
a result, the following matrix is obtained.
\begin{equation}
\label{1}
\begin{bmatrix}
0& 0& 0& 0& 0& 0& 0& 0& 0& 0& 0& 0& 0& 0 \\
0& 0& 0& 0& 0& 0& 0& 0& 0& 0& 0& 0& 0& 0 \\
0& 0& 0& 0& 0& 0& 0& 0& 0& 0& 0& 0& 0& 0 \\
0& 0& 0& 0& 0& 0& 0& 0& 0& 0& 0& 0& 0& 0 \\
0& 0& 0& 0& 1& 1& 1& 1& 1& 1& 1& 1& 0& 0 \\
0& 0& 0& 0& 1& 1& 1& 1& 1& 1& 1& 1& 0& 0 \\
0& 0& 0& 0& 1& 1& 1& 1& 1& 1& 1& 1& 0& 0 \\
0& 0& 0& 0& 1& 1& 1& 1& 1& 1& 1& 1& 0& 0 \\
0& 0& 0& 0& 1& 1& 1& 1& 1& 1& 1& 1& 0& 0 \\
0& 0& 0& 0& 1& 1& 1& 1& 1& 1& 1& 1& 0& 0 \\
0& 0& 0& 0& 1& 1& 1& 1& 1& 1& 1& 1& 0& 0 \\
0& 0& 0& 0& 1& 1& 1& 1& 1& 1& 1& 1& 0& 0 \\
0& 0& 0& 0& 0& 0& 0& 0& 0& 0& 0& 0& 0& 0 \\
0& 0& 0& 0& 0& 0& 0& 0& 0& 0& 0& 0& 0& 0
\end{bmatrix}
\end{equation}
Now dividing the image into a $(7 \times 7)$ grid. Similarly, the
corresponding $(7 \times 7)$ matrix can be defined to identify the
object center point as follows.
\begin{equation}
\label{2}
\begin{bmatrix}
0 &  0 &  0 &  0 &  0 &  0 &  0 \\
0 &  0 &  0 &  0 &  0 &  0 &  0 \\
0 &  0 &  0 &  0 &  0 &  0 &  0 \\
0 &  0 &  0 &  0 &  0 &  0 &  0 \\
0 &  0 &  0 &  0 &  1 &  0 &  0 \\
0 &  0 &  0 &  0 &  0 &  0 &  0 \\
0 &  0 &  0 &  0 &  0 &  0 &  0
\end{bmatrix}
\end{equation}

\textcolor{black}{A point to ponder  is, to obtain the ROI, one may utilize other matrices (e.g. $(28 \times 28)$ or $(16 \times 16)$) rather than the exact $(14 \times 14)$ matrix. The important thing is to utilize the output of a convolutional layer in Fig. \ref{fig1} which results in a better outcome. Taking this structure into consideration, the corresponding proper choice is the $(14 \times 14)$ matrix. This is because of two reasons; First, down-sampling
yields richer information, and second, more convolutional layers result in a more flexible structure. Furthermore, to employ such a matrix
in determining the ROI, post-processing is necessary which makes a
smaller matrix with richer features, a better choice in terms of
both the accuracy and the speed. On the other hand, the use of the second
$(7 \times 7)$ matrix is important in preventing failures in certain
situations. This point is covered in the experimental results
section in more detail. This matrix helps to get more robust results,
and furthermore, it produces another gradient flow which makes the
$(14 \times 14)$ matrix more oriented towards the object location. The same point mentioned about utilizing other matrices rather than the exact $(14 \times 14)$ is valid about the $(7 \times 7)$ matrix, as well.}
Taking these two matrices as ground truth, the following loss
function is defined:

\begin{equation}
\label{3} L = \alpha_1 (\sum(\gamma^{2(\hat{p}_1 - p_1)})) +
\alpha_2 (\sum(\gamma^{2(\hat{p}_2 - p_2)}))
\end{equation}
where $p_1$ and $p_2$ are the true values of $(14 \times 14)$ and
$(7 \times 7)$ matrices, and $\hat{p}_1$ and $\hat{p}_2$ are their
estimats obtained by the CNN, respectively.

Remark 3: $\alpha_1$ and $\alpha_2$  are the coefficients used to
compensate for the bias in each term, and $\gamma$ is a positive
value less than one ($ 0< \gamma < 1 $).

Remark 4: Note that the matrices $\hat{p}_1$ and $\hat{p}_2$ are
obtained by using an average over the feature-maps
followed by the "softmax" function.

As it can be seen, equation (\ref{3}) contains two terms, each one
corresponding to the relating matrix. The first term is responsible
for determining the object's ROI. This will result the CNN to
generate a matrix in which lae larger component of the matrix
correspond to higher probability of existence of a part of object in
that component. The second term is used for identifying the object
center point in which the $(7 \times 7)$ matrix is utilized. Note
that, utilizing few layers to get the $(7 \times 7)$ matrix from
the $(14 \times 14)$ matrix by down-sampling effectively reduces the
number of probable anchor boxes when this CNN is employed in a
two-stage detector.

Remark 5: Comparing the proposed loss function to commonly used mean
square error one, it has the advantage of magnifying the importance of
the error in the components related to the object location. By this
means, the necessity of finding which components may contain a part of an
object is well addressed.

\subsection{The Dataset}
At this stage, let us elaborate on the proper dataset used to train
the proposed network. Reconsider loss function (\ref{3}),where for
each image, the two $(14 \times 14)$ and $(7 \times 7)$ matrices
shall be produced. Taking into consideration the fact that the
existing annotations often contain bounding boxes, it is very simple
to get the required matrices via them. Since a single object
tracking application is taken to further illustrate the power of
both the proposed training approach and the flexible structure, a
dataset called "NFS"~\cite{NFS} is used, which is known as an
object tracking benchmark. Necessary modifications are performed on
this dataset to produce the required data.

\begin{figure}[]
    \centering
    \includegraphics[width=4in]{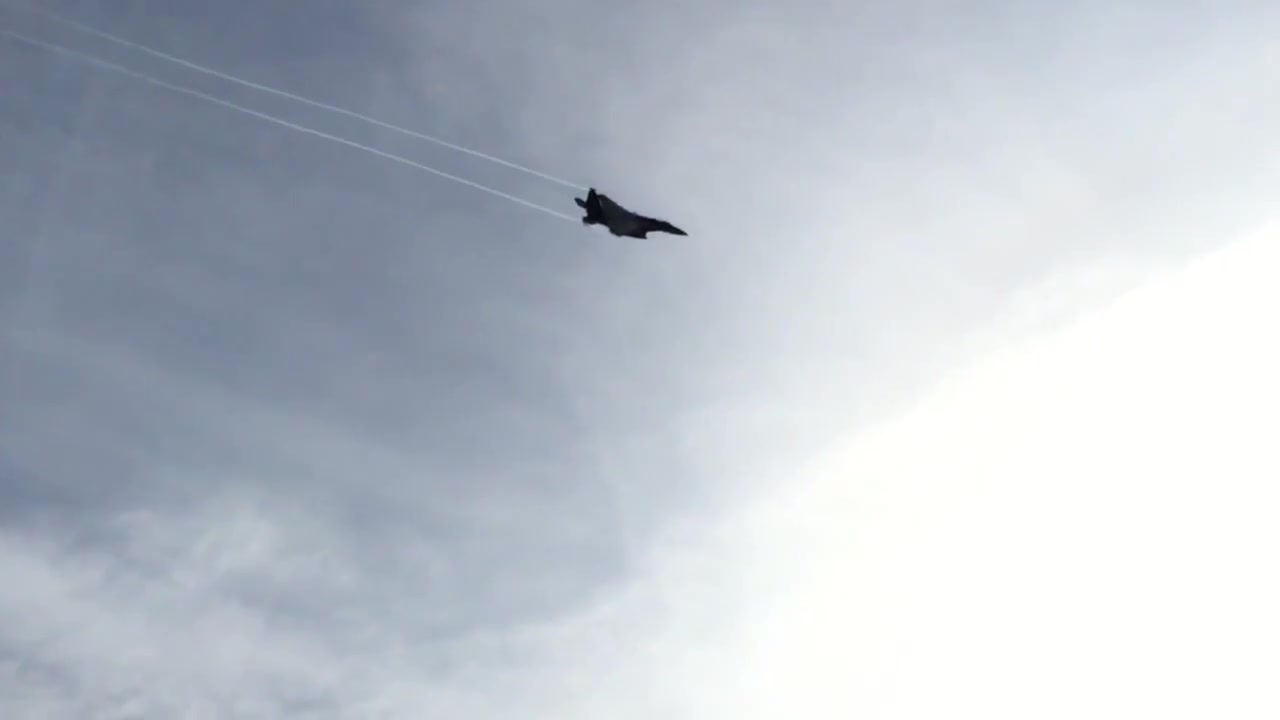}
    \caption{\small{A sample image from the NFS dataset}}
    \label{fig2_}
\end{figure}

\begin{figure}[]
    \centering
    \includegraphics[width=2in]{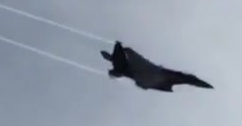}
    \caption{\small{The cropped part which is used as the CNN input for the tracking application}}
    \label{fig3}
\end{figure}

\begin{figure}[h]
    \centering
    \subfloat[]{\includegraphics[width=1in]{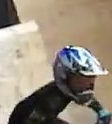}
        \label{fig4_a}}
    \subfloat[]{\includegraphics[width=1in]{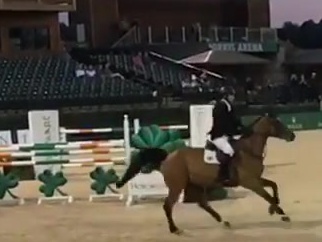}
        \label{fig4_aa}}
    \subfloat[]{\includegraphics[width=1in]{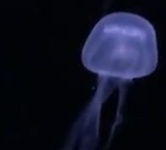}
        \label{fig4_b}}\\
    \subfloat[]{\includegraphics[width=1in]{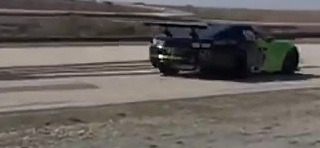}
        \label{fig4_c}}
    \subfloat[]{\includegraphics[width=1in]{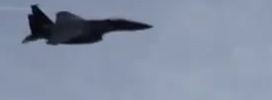}
        \label{fig4_bb}}
    \subfloat[]{\includegraphics[width=1in]{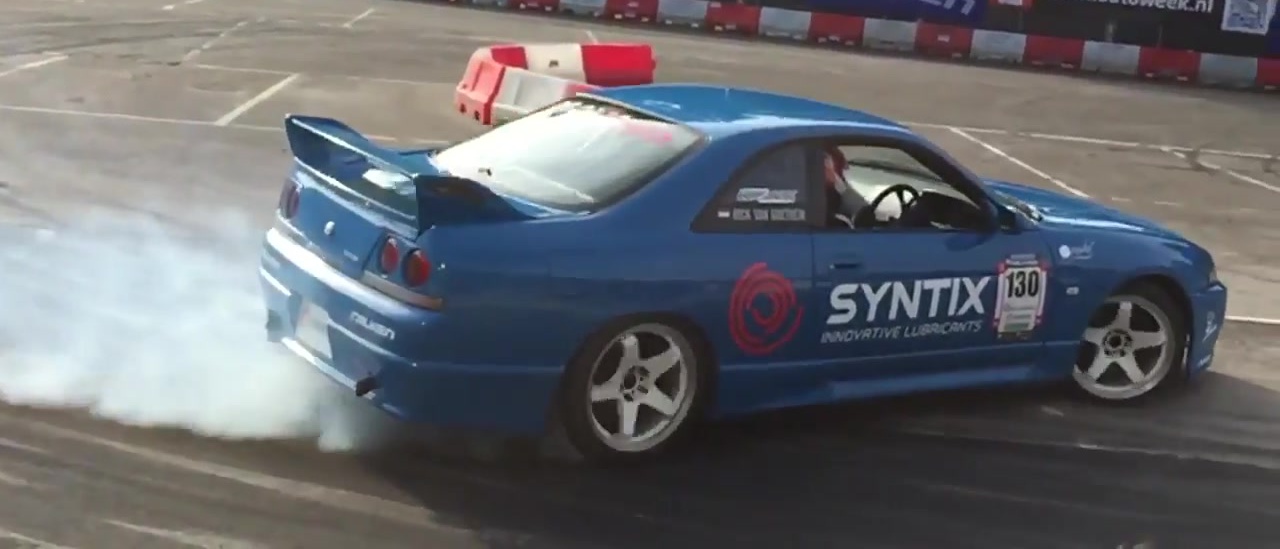}
        \label{fig4_d}}\\
    \subfloat[]{\includegraphics[width=1in]{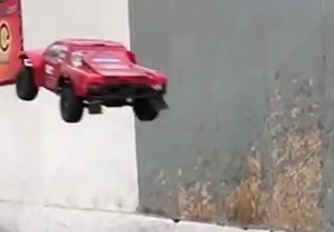}
        \label{fig4_e}}
    \subfloat[]{\includegraphics[width=1in]{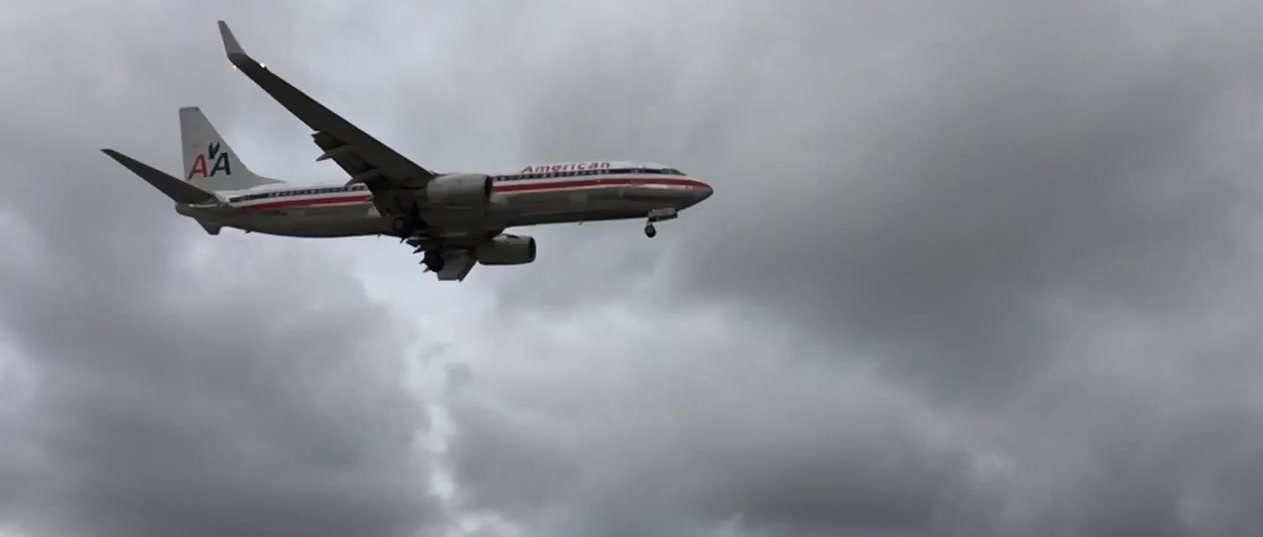}
        \label{fig4_cc}}
    \subfloat[]{\includegraphics[width=1in]{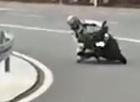}
        \label{fig4_f}}\\
    \caption{Some representative samples of the produced dataset }
    \label{fig4}
\end{figure}

To describe the process, consider Fig.~\ref{fig2_} as the main
image.
Since in the tracking application at the first frame the user
determines the requested object to be tracked, it can be assumed
that the approximate object location is given. Therefore, it is
rational to process just a part of the image rather than the whole
image. In this regard, Fig. \ref{fig3} indicates the relevant part
of the image which will be used as input to train the network. Note
that, here the object may be located in the image center, however,
in the next frame, as there is no information about where the object
is, the same part of the image is a proper choice to be used.
Apparently, this time the object is not located at the center since
it has moved from its previous location. Hence, in order to be
capable of detecting the object in any location of the input image,
it is better to randomly locate objects in various places. To
further illustrate this important point, in Fig. \ref{fig4} some
representative samples of the produced dataset is shown. It can be
seen that the objects are not located at the center,  different
object types are considered, while the images have various sizes.

Remark 6: A point to ponder is that the annotations of the NFS
dataset have to be changed according to the new images indicated in
Fig.~\ref{fig4}.

\section{Experimental Results}
In this section, the results of training the proposed structure
illustrated in Fig.~\ref{fig1} on the produced dataset are reported.
The training approach is the same as suggested in the previous
section in which $\alpha_1=10^{-5}$, $\alpha_2=234$, and
$\gamma=0.0001$ are used. The optimizer used to train the CNN is the
stochastic gradient descent (SGD) with $0.001$ learning rate, and
the CNN has been trained for about $65$ epochs. Furthermore, to
avoid having any bias in the dataset, about $8700$ frames are
selected in which there are approximately $500$ images for each
object type. Considering different object types, the input images
will have various sizes activating different branches of the
proposed structure. In what follows, first, some test images with
different sizes will be processed and the CNN outputs are
investigated.
\textcolor{black}{Second, a study is reported on the possibility of using higher resolution ROI matrices such as $(28 \times 28)$}.
Then, the trained CNN is employed as a single object tracker
considering three scenarios. Several challenges in the
implementation are addressed. To evaluate the proposed
tracker, the results are compared to some commonly used trackers
such as GOTURN and other OpenCV trackers with respect to a common
measure. \textcolor{black}{Finally, to further demonstrate the proposed approach capabilities, the tracker is trained on another dataset called LaSOT\cite{lasot} and it is compared to a recently presented tracker to analyze its performance in more detail.}

\subsection{Results on Images with Various Sizes}
\begin{figure}[t]
    \centering
    \subfloat[]{\includegraphics[width=2.3in]{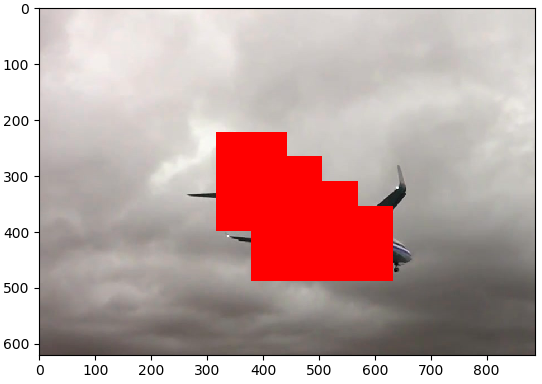}
        \label{fig6_a}}
    \subfloat[]{\includegraphics[width=2.3in]{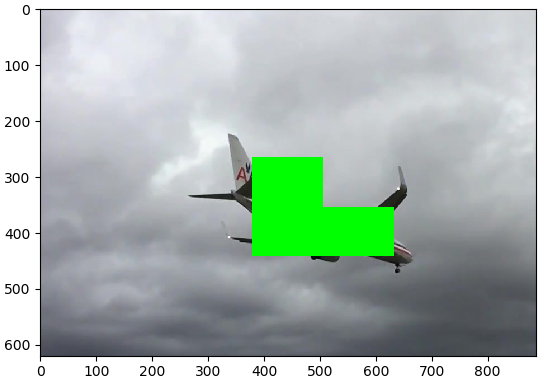}
        \label{fig6_b}}\\
    \subfloat[]{\includegraphics[width=2.3in]{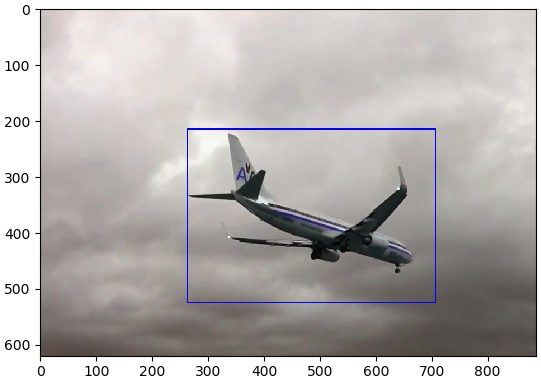}
        \label{fig6_c}}\\
    \caption{Visual outcomes of the ROI matrix alongside the center matrix, and the fitted bounding box}
    \label{fig6}
\end{figure}

First, let's investigate the performance and the details of the
algorithm through an image. Thereafter, various image sizes are
processed to further demonstrate the applicability of both the
suggested structure and the proposed training method. Reconsider
Fig.~\ref{fig2} which has been classified to the $(448 \times 448)$
size. Taking this image as the CNN input, the output for the $(14
\times 14)$ matrix is obtained as:

\scriptsize{\begin{equation}
\label{4}
{\begin{bmatrix}
0.06 & 0.07 & 0.07 & 0.07 & 0.07 & 0.07 & 0.07 & 0.07 & 0.07 & 0.07 & 0.07 & 0.07 & 0.07 & 0.06  \\
0.06 & 0.07 & 0.07 & 0.07 & 0.07 & 0.07 & 0.07 & 0.07 & 0.07 & 0.07 & 0.07 & 0.07 & 0.07 & 0.06 \\
0.06 & 0.07 & 0.07 & 0.07 & 0.07 & 0.07
& 0.07 & 0.07 & 0.07 & 0.07 & 0.07 & 0.07 & 0.07 & 0.06 \\
0.06 & 0.06 & 0.07 & 0.07 & 0.07 & 0.07
& 0.07 & 0.07 & 0.07 & 0.07 & 0.07 & 0.07 & 0.06 & 0.06 \\
0.06 & 0.06 & 0.06 & 0.07 & 0.07 & 0.08
&0.08 & 0.07 & 0.06 & 0.06 & 0.06 & 0.06
&0.06 & 0.06 \\
0.05 & 0.05 & 0.06 & 0.06 & 0.07 & 0.09 & 0.09 & 0.08 & 0.07 & 0.07 & 0.06 & 0.06 & 0.06 & 0.05 \\
0.05  & 0.05 & 0.05 & 0.06 & 0.07 & 0.09
& 0.10 & 0.09 & 0.08 & 0.07 & 0.07 & 0.06 & 0.05 & 0.05 \\
0.04 & 0.04 & 0.04 & 0.05 & 0.07 & 0.09
& 0.11 & 0.10 & 0.09 & 0.08 & 0.07 & 0.06 & 0.04 & 0.04 \\
0.04 & 0.04 & 0.04 & 0.052 & 0.06 & 0.09
& 0.11 & 0.11 & 0.10 & 0.09 & 0.07 & 0.05 & 0.04 & 0.04 \\
0.04 & 0.04 & 0.04 & 0.05 & 0.06 & 0.08 & 0.10 & 0.11 & 0.11 & 0.09 & 0.07 & 0.06 & 0.05 & 0.04 \\
0.05 & 0.05 & 0.05 & 0.05 & 0.06 & 0.07
& 0.09 & 0.101 & 0.10 & 0.092 & 0.07 & 0.06
& 0.05 & 0.05 \\
0.06 & 0.06 & 0.06 & 0.06 & 0.06 & 0.07
& 0.07 & 0.08 & 0.08 & 0.08 & 0.07 & 0.06 & 0.06 & 0.05 \\
0.06 & 0.07 & 0.07 & 0.07 & 0.06 & 0.07
& 0.07 & 0.07 & 0.07  & 0.07 & 0.07 & 0.07
& 0.06 & 0.06 \\
0.07 & 0.07 & 0.07 & 0.07 & 0.07 & 0.07
& 0.07 & 0.07 & 0.07 & 0.07 & 0.07 & 0.07 & 0.06 & 0.06
\end{bmatrix}}
\end{equation}}
\normalsize{It is easy to identify a proper threshold by solving an
optimization problem over the test dataset to maximize the overlap
between the object location and the components in the $(14 \times
14)$ matrix. Here, the threshold is set to $0.09$ by which the ROI
will be accurately determined. Fig.~\ref{fig6_a} indicates a visual
outcome of using this threshold on the $(14 \times 14)$ matrix in
which the red-colored parts represent the matrix components with
values higher than $0.09$. As it can be seen, the result is
promising and the location of the object has been precisely
identified. Now, let's check the $(7 \times 7)$ matrix as it is
responsible for presenting probable center point candidates:}

\scriptsize{ \begin{equation} \label{5} {\begin{bmatrix}
    0.14285 & 0.14285 & 0.14285 & 0.14285 & 0.14285 & 0.14285 & 0.14285 \\
    0.14285 & 0.14285 & 0.14285 & 0.14285 & 0.14285 &  0.14285 & 0.14285 \\
    0.14282 & 0.14282 & 0.14282 & 0.14303 & 0.14282 &  0.14282 & 0.14282 \\
    0.11155 & 0.11155 & 0.14936 & 0.22829 & 0.17611 &  0.11155 & 0.11155 \\
    0.10102 & 0.10102 & 0.13884 & 0.24611 & 0.2102 &   0.10171 & 0.10102 \\
    0.13189 & 0.13189 & 0.13189 & 0.17523 & 0.16529 &  0.13189 & 0.13189 \\
    0.14285 & 0.14285 & 0.14285 & 0.14285 & 0.14285 &  0.14285 & 0.14285
    \end{bmatrix}}
\end{equation}}
\normalsize{Similarly, it is easy to find an appropriate threshold
which is set as $0.18$ in here. Fig.~\ref{fig6_b} illustrates the
outcome of using this threshold, in which, the matrix components
higher than $0.18$ are displayed with the green-colored pixels.
Utilizing these two matrices, the ROI along with the probable center
points of the object are found. However, the problem of the
computational cost is predominant. Thus, the easiest way of
searching for the values higher than the determined thresholds might
not be proper computational-wise. In this regard, a specific
function \texttt{"max pool with argmax()"} is used on the $(14
\times 14)$ matrix to find the larger values than the threshold and
their arguments. Thereby, instead of searching through $(14 \times
14)$ components, values higher than the threshold are found quite
quickly. Having these arguments, it is possible to modify the
bounding box scale. The higher the values over the threshold, the
higher the scale. Next, for the $(7 \times 7)$ matrix the same
approach is used to identify the candidates, and the mean value of
them will be utilized as the final center point for the bounding
box.}

\begin{figure}[]
    \centering
     \subfloat[]{\includegraphics[width=0.8in,height=0.8in]{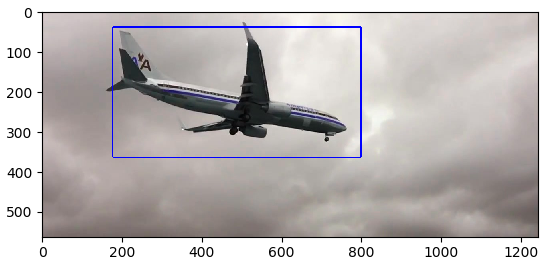}
        \label{fig7_1}}
    \subfloat[]{\includegraphics[width=0.8in,height=0.8in]{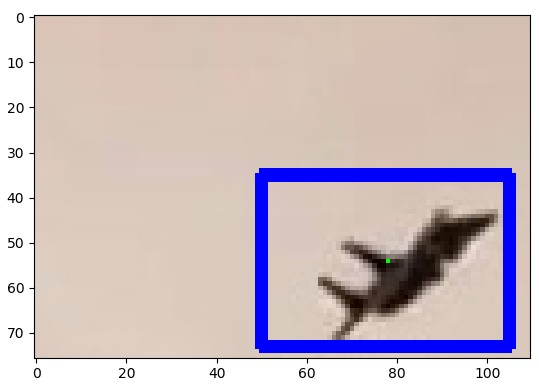}
        \label{fig7_2}}
    \subfloat[]{\includegraphics[width=0.8in,height=0.8in]{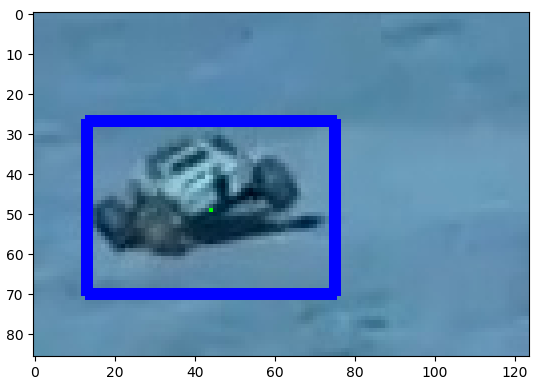}
        \label{fig7_3}}
    \subfloat[]{\includegraphics[width=0.8in,height=0.8in]{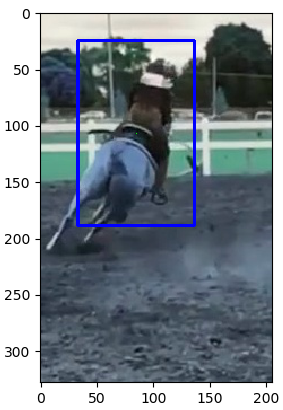}
        \label{fig7_4}}
    \subfloat[]{\includegraphics[width=0.8in,height=0.8in]{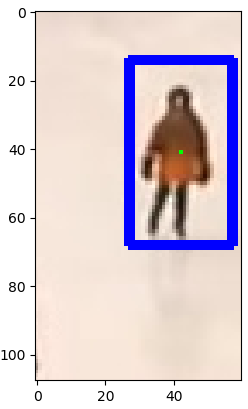}
        \label{fig7_5}} \\
    \subfloat[]{\includegraphics[width=0.8in,height=0.8in]{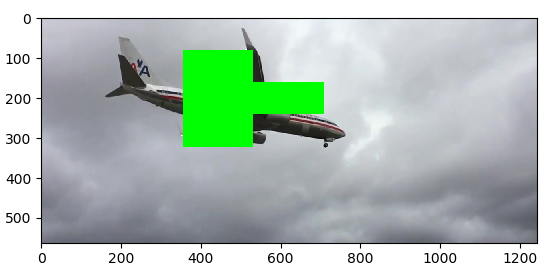}
        \label{fig7_6}}
    \subfloat[]{\includegraphics[width=0.8in,height=0.8in]{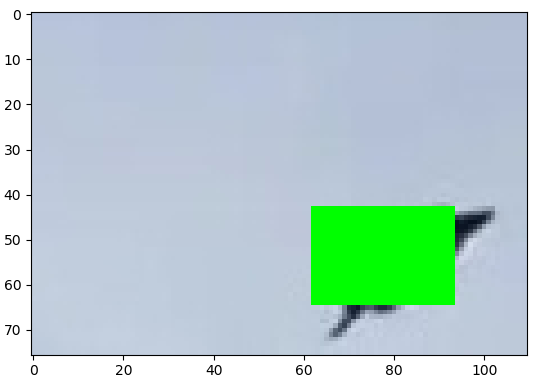}
        \label{fig7_7}}
    \subfloat[]{\includegraphics[width=0.8in,height=0.8in]{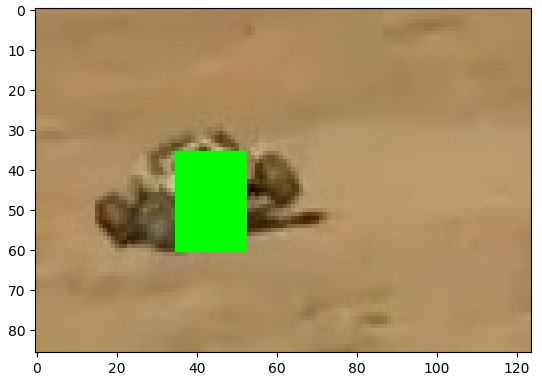}
        \label{fig7_8}}
    \subfloat[]{\includegraphics[width=0.8in,height=0.8in]{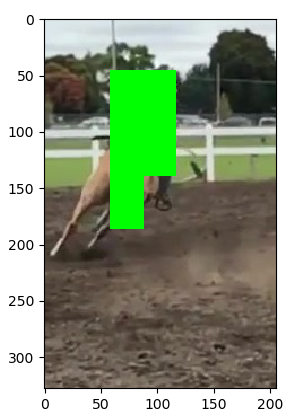}
        \label{fig7_9}}
    \subfloat[]{\includegraphics[width=0.8in,height=0.8in]{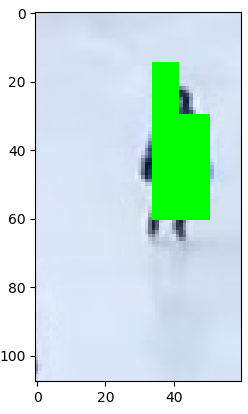}
        \label{fig7_10}}\\
    \subfloat[]{\includegraphics[width=0.8in,height=0.8in]{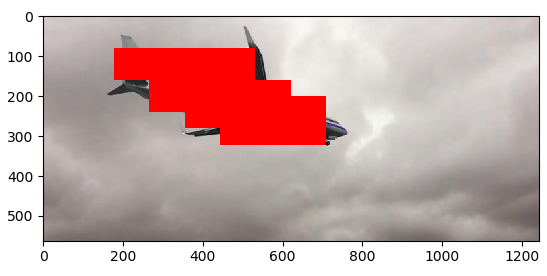}
        \label{fig7_11}}
    \subfloat[]{\includegraphics[width=0.8in,height=0.8in]{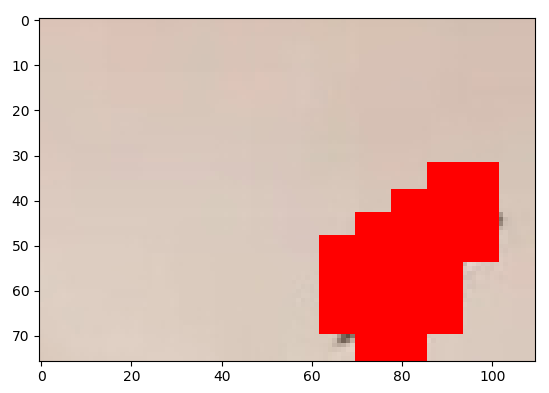}
        \label{fig7_12}}
    \subfloat[]{\includegraphics[width=0.8in,height=0.8in]{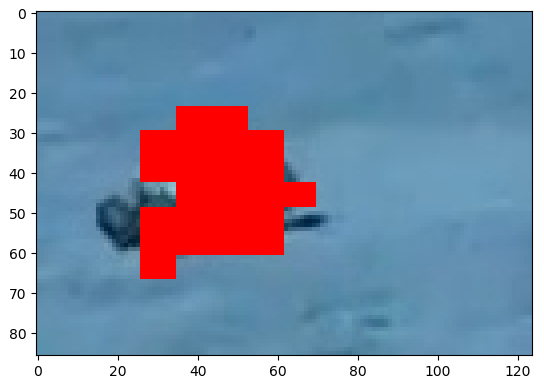}
        \label{fig7_13}}
    \subfloat[]{\includegraphics[width=0.8in,height=0.8in]{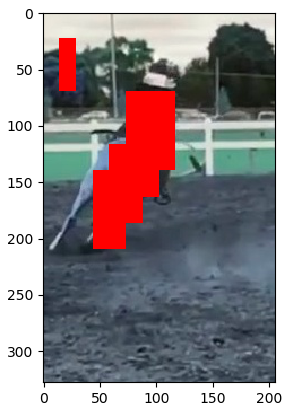}
        \label{fig7_14}}
    \subfloat[]{\includegraphics[width=0.8in,height=0.8in]{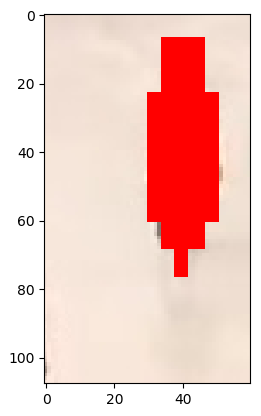}
        \label{fig7_15}}\\
    \caption{Some samples of the test results.}
    \label{fig7}
\end{figure}

\begin{figure}[]
    \centering
    \includegraphics[width=5in]{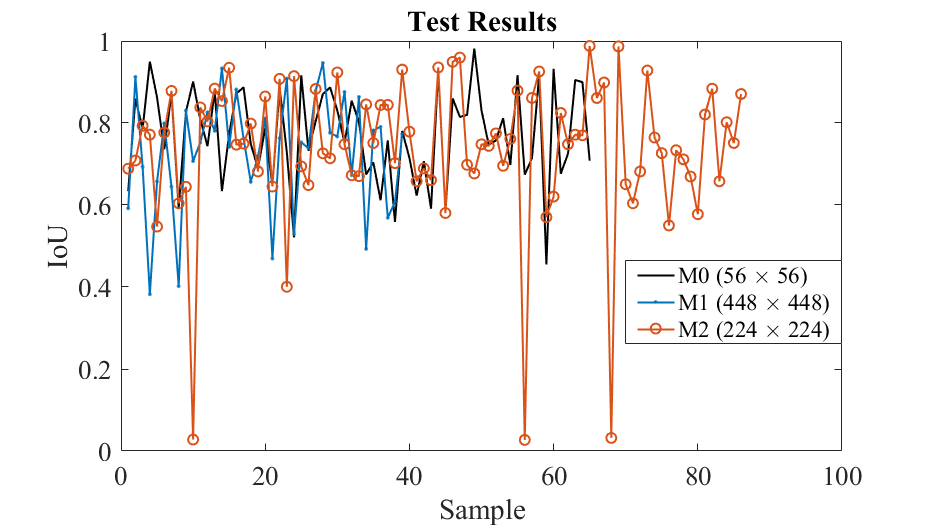}
    \caption{\small{IoU for several test images considering various sizes.}}
    \label{fig8}
\end{figure}

In object tracking the target is to locate the object at each frame.
Having some information from the initial frame, it is possible to
modify and shift the initial bounding box by the two $(14 \times
14)$ and $(7 \times 7)$ matrices. Of course, there are other ways to
implement such an algorithm; however, since the speed is of high
importance, it has been tried to use nearly the fastest approach.
Fig. \ref{fig6_c} indicates the most probable bounding box for the
object. To demonstrate the applicability of the proposed structure,
some other images with various sizes are tested and the results are
reported in Fig.~\ref{fig7}. In this figure three rows are presented
which illustrate the visual outcomes of the ROI matrix, the center
point matrix, and the most probable bounding boxes, respectively.
Among over $350$ test images the average intersection over union
(IoU) is about $0.75$ which verifies the effectiveness of the
proposed approach. Furthermore, Fig.~\ref{fig8} shows the IoU
obtained from processing lots of images with various sizes. In this
figure, three lines represent the results of each of the three
mentioned models. As it is seen, results are promising and the
suggested CNN has uniformly suitable performance on various image
sizes.


\textcolor{black}{
\subsection{A Study on a Higher Resolution ROI Matrix}
In this part of the article, the possibility of using a higher
resolution ROI matrix is studied. In this respect, the proposed
architecture is modified and reduced with less convolutional layers
to output a $(28 \times 28)$ matrix. Then, it is trained on the NFS
dataset for $100$ epochs. Note, in the reduced structure the
convolutional layers are reduced and the problem complexity on the
other hand, has been increased due to the higher resolution of the
input image. In order to have a fair comparison between the two
results, a metric is used on a test set which counts the $L1$ norm
of the error between an output and the given ground truth.
Furthermore, since the proposed architecture has three branches,
there are three outcomes corresponding to the three models in the report. To get better
comparison the $(14 \times 14)$ matrix is resized to a $(28 \times28)$
one, such that it could be directly compared to that of the other
approach that uses the $(28 \times 28)$ matrix.
}
\begin{figure}[]
    \centering
    \includegraphics[width=4.5in]{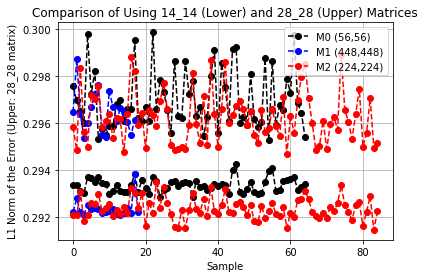}
    \caption{Results of employing the presented approach and a reduced architecture with higher resolution.}
    \label{fig_higher_res}
\end{figure}
\textcolor{black}{
By comparing the results given in Fig. \ref{fig_higher_res}, it is
clear that the outcomes of utilizing the presented architecture are
more reliable in all the three models while the accuracy is higher.
As a consequence, it is readily possible to employ a higher
resolution matrix. However, considering all the details presented in
this section, it is proposed to use the $(14 \times 14)$ matrix in
practice. }

\subsection{Results of the Proposed Tracker}
Now that the performance of the CNN on individual images is
verified, some tracking scenarios are presented by which a
comparative study over eight other trackers is reported, and the
performance of the proposed tracking algorithm is evaluated. The
purpose of considering these scenarios is to evaluate the results of
the suggested approach in both estimating center points and
identifying the object scale. The IoU at each frame is considered as
a metric of the tracker performance. \textcolor{black}{Note that,
the only limitation here in tracking, is to always have the object
in the camera field of view.} Before reporting the results, a brief
description of each tracker is presented to further get insight to
their characteristics. The first tracker used is called Boosting,
one of the oldest trackers which has a significant performance in
detections; however, since it is too slow, it can be utilized only
as a tool for comparing the other designed trackers \cite{BOOSTING}.
Multi instance learning (MIL) tracker leads to a robust tracking
compared to that of the other traditional trackers. Although this
tracker is more accurate than the Boosting tracker, it faces more
problems to accurately report the failures~\cite{MIL}. The other
tracker is the kernelized correlation filter (KCF) which operates
faster than MIL and Boosting trackers but it may not handle the
occlusion~\cite{KCF}. Dealing with tracking the occluded objects,
channel and spatial reliability tracker (CSRT) operates more
efficiently, with the expense of being slower in
computation~\cite{CSRT}. Medianflow tracker has been proposed to
particularly track small and low-contrast objects~\cite{MEDIANFLOW}.
However, it fails to track fast-moving objects and also the objects
with quick changes in the appearance. This tracker, however, reports
the failures very accurately. It also tracks the object in both
forward and backward directions which reduces the error
significantly. In the tracking-learning-detection (TLD), the tracker
follows the detected object at each frame, the detector localize
observed features and the learning module is responsible to reduce
the error in the future frames~\cite{TLD}. The high speed MOSSE
tracker, is reported to be robust against variations in lighting,
scale, pose, and nonrigid deformations~\cite{MOSSE}. Finally, the
only CNN based tracker here, GOTURN~\cite{GOTURN} utilizes two
frames to track the object while employing a special structure.

\begin{figure}[]
    \centering
    \includegraphics[width=4.5in]{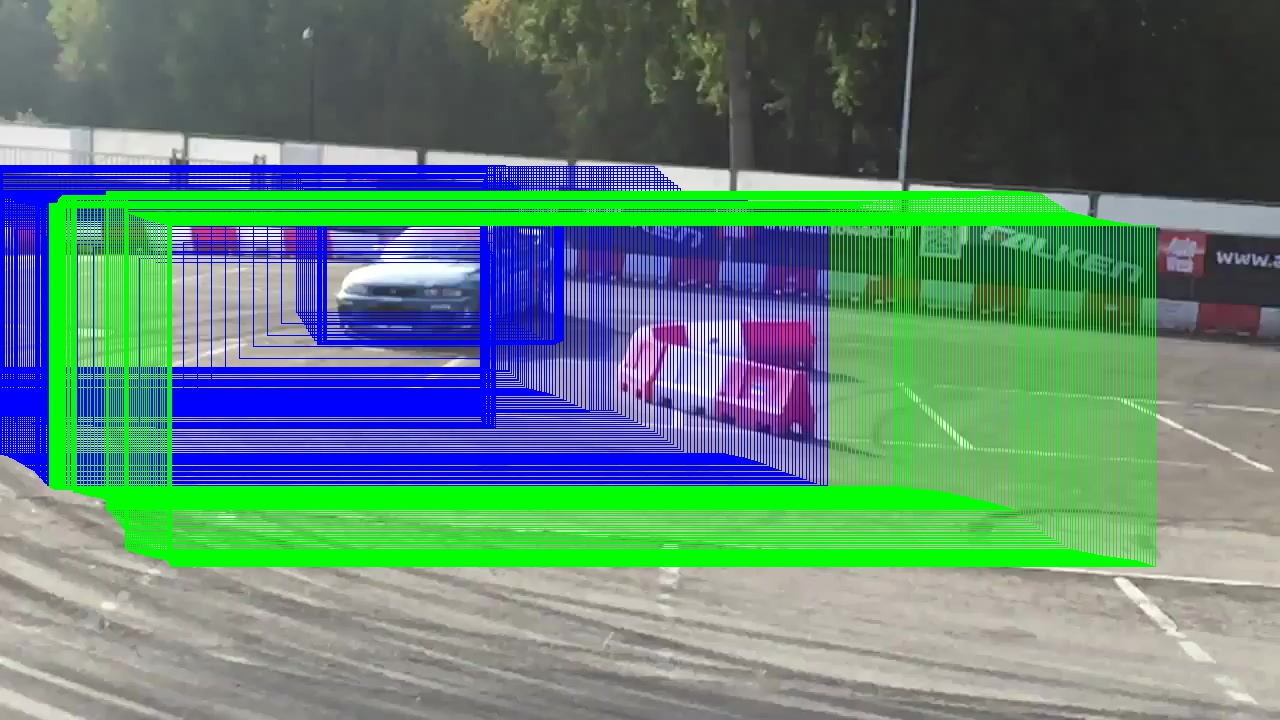}
    \caption{\small{The first scenario (A drifting car). The colored bounding boxes indicate the motion trajectory.}}
    \label{fig9}
\end{figure}

\begin{figure}[]
    \centering
    \subfloat[]{\includegraphics[width=2.3in]{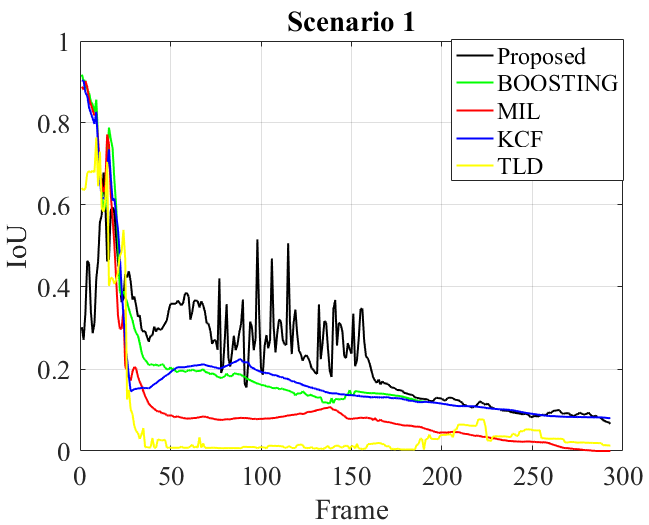}
        \label{fig10_a}}
    \subfloat[]{\includegraphics[width=2.3in]{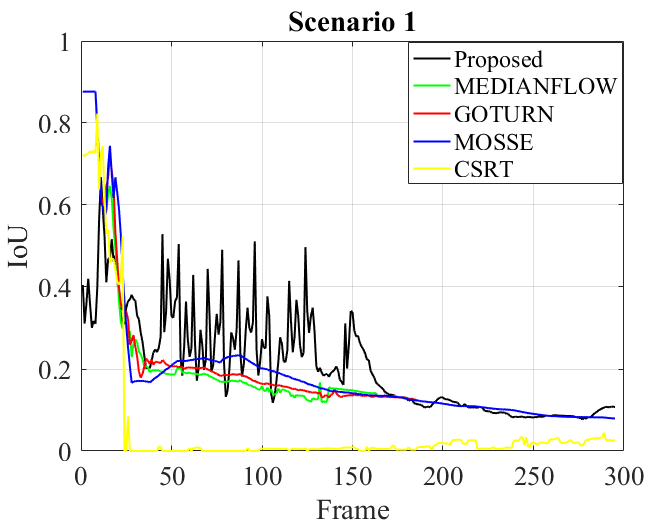}
        \label{fig10_b}}\\
    \caption{Results of the trackers for the first scenario.
    Each tracker's outcomes are given with a specific colored-line
    and the proposed tracker has been shown by the black line in
    both figures.}
    \label{fig10}
\end{figure}

Fig.~\ref{fig9} illustrates the first experiment scenario in which
there is a car drifting over the road and the challenge is to keep
the IoU high, while the object is becoming larger in the image. In
all the reported scenarios the moving trajectory of the object is
illustrated by the colored bounding boxes. To compare the proposed
tracker with the others, the IoU is determined at each frame, and
the results are reported in two diagrams to quantitatively compare
the performance of the trackers. Fig.~\ref{fig10} indicates the
results obtained from various trackers in this experiment. As it is
seen in this figure, despite the low IoU at the beginning, the
proposed tracker has almost constant performance regardless of the
object changing size. On the other hand, other trackers perform
poorly as the object size begins to change quickly. Note that the
fluctuations seen in the IoU of the proposed approach, indicate that
the tracker is trying to modify the bounding box to keep the IoU as
high as possible.

\begin{figure}[]
    \centering
    \includegraphics[width=4.5in]{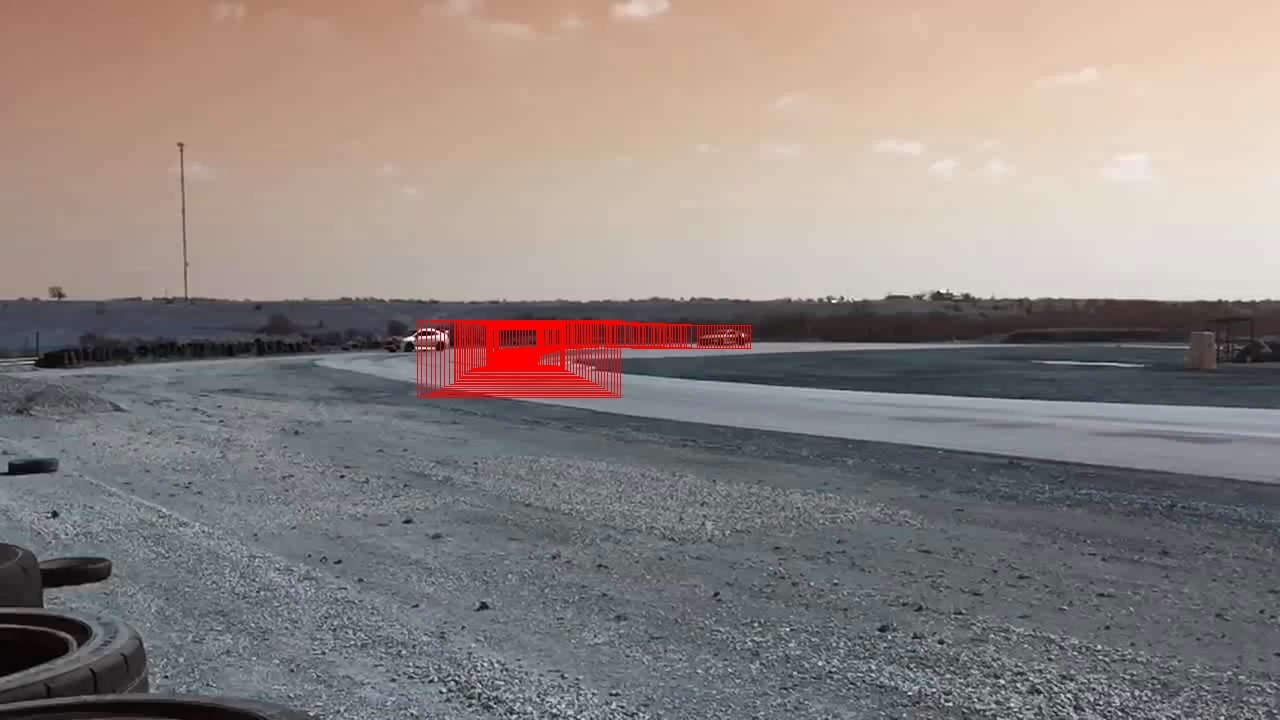}
    \caption{\small{The second scenario (a far away car).}}
    \label{fig11}
\end{figure}

\begin{figure}[]
    \centering
    \subfloat[]{\includegraphics[width=2.45in]{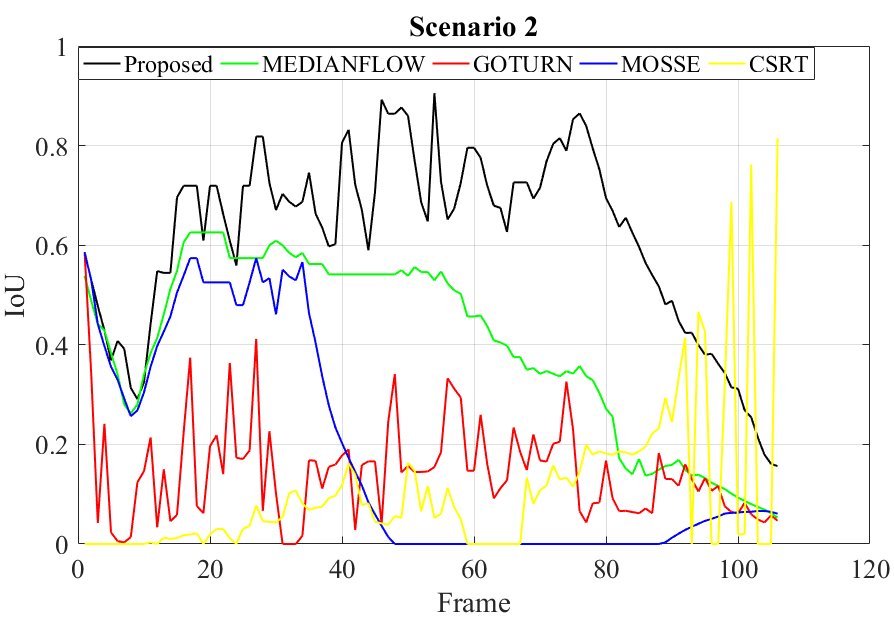}
        \label{fig12_a}}
    \subfloat[]{\includegraphics[width=2.3in]{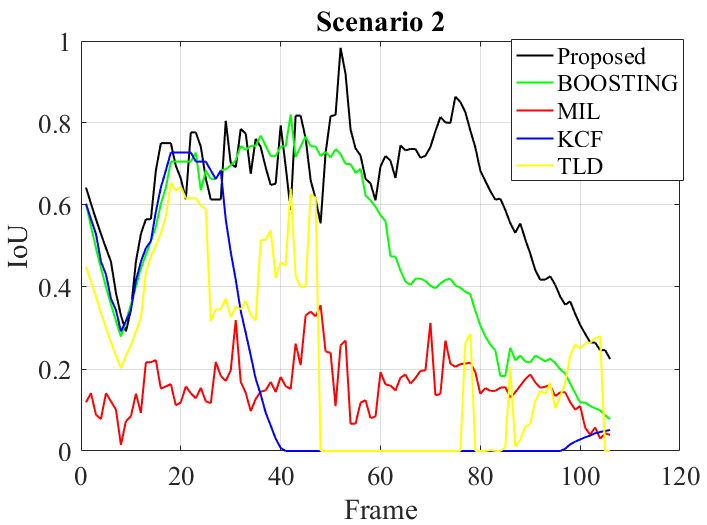}
        \label{fig12_b}}\\
    \caption{Results of the trackers for the second scenario}
    \label{fig12}
\end{figure}

The second experiment scenario is shown in Fig.~\ref{fig11} in which
a car is moving towards the camera while due to its far distance it
has a relatively small size in the image. Here the challenge is
different from the previous scenario, in the sense that the object
size is much smaller while there is a shift in the object center
point to be estimated precisely. This is due to the fact that a
small error in the center point may result in poor performance or
even losing the object features in practice. Note that in this
scenario the camera moves as well. The results of using the trackers
are reported in Fig.~\ref{fig12}. As shown in Fig.~\ref{fig12_a},
the performance of the proposed tracker is almost better everywhere.
In here the GOTURN tracker has failed to perform well, and compared
to the conventional methods, MEDIANFLOW tracker performs the best.
This result deserves more attention since it highlights the fact
that CNN based trackers such as GOTURN may perform poor in scenarios
like this while the MEDIANFLOW could possibly yield a better
outcome. However, the proposed tracker has performed significantly
better than the best of other trackers.

\begin{figure}[]
    \centering
    \includegraphics[width=4.5in]{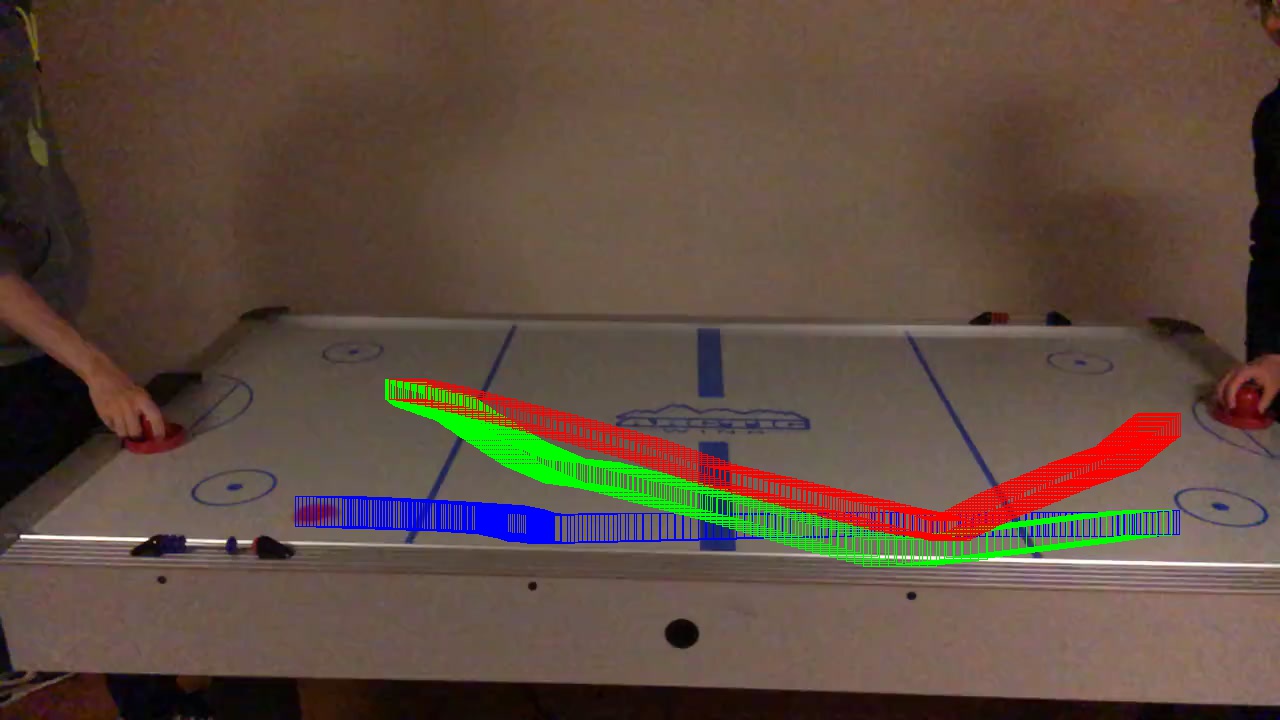}
    \caption{\small{The third scenario (a red disk).}}
    \label{fig13}
\end{figure}

Finally,  in the last experiment scenario, the purpose is to
evaluate the proposed tracker in long term tracking while the object
center point significantly varies in both directions. As it can be
seen in Fig.~\ref{fig13}, a red disk moves through the blue, the
green, and the red bounding boxes, respectively. These boxes
illustrate the location of the disk in the next frames. Furthermore,
the color of the boxes is changed whenever it comes to a hit. Note
that the game tools are quite similar to the red disk which may
yield failure when it comes to a hit. This will help to explain why
both the $(14 \times 14)$ and $(7 \times 7)$ matrices are needed. In
fact, considering the information obtained from the $(14 \times 14)$
matrix, one may think that utilizing this matrix, it is possible to
estimate the center point, as well. However, in cases like such
experiment, when the tracked object get close to another similar
object in the same region, using only the $(14 \times 14)$ matrix
will result in failure. \textcolor{black}{This is because utilizing
the  single $(14 \times 14)$ matrix may result in false outcomes
since there are some other objects or surfaces in the environment
which may seem just like a part of the object. As a result,
employing this matrix to obtain the center point will result in
errors in situations like this.}
\begin{figure}[]
    \centering
    \subfloat[]{\includegraphics[width=2.3in]{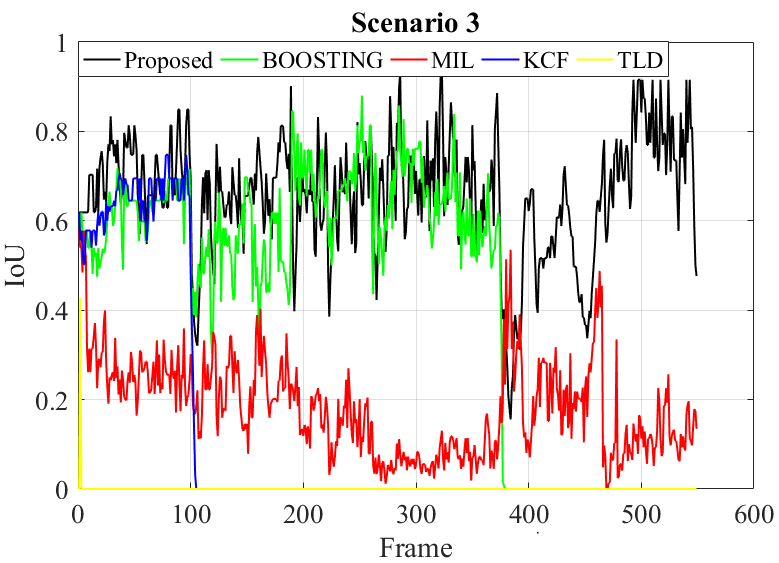}
        \label{fig14_a}}
    \subfloat[]{\includegraphics[width=2.3in]{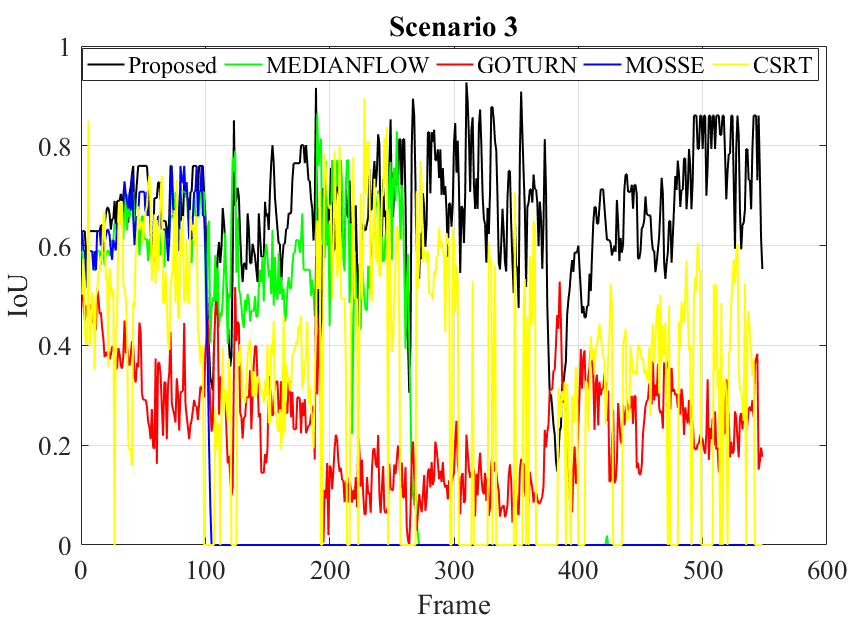}
        \label{fig14_b}}\\
    \caption{Results of the trackers for the third scenario}
    \label{fig14}
\end{figure}

Fig.~\ref{fig14} indicates the performance of the trackers in
detail. In Fig. \ref{fig14_a} it can be seen that although the
BOOSTING tracker has performed well before its failure it could not
distinguish the red disk with the red game tool at the second
contact point. Regarding the results, both the proposed tracker and
the MIL tracker has accomplished the task. However, the results of
the proposed tracker are far better in terms of IoU measure.
Fig.~\ref{fig14_b} illustrates the functionality of the other
trackers. As it is seen in this fugure, the MEDIANFLOW tracker has
performed just like the BOOSTING, however, it has failed at the same
occasion. In this result, there are three trackers capable of
tracking the object during the whole experiment, while the proposed
tracker has the best performance. Although the CSRT tracker did not
fail and it has accomplished the task, it has very large
fluctuations that downgrade its functionality, and maybe the
GOTURN's performance is more desirable in practice.

Notice that one important aspect of a tracker is its speed.
Considering the light structure of the proposed tracker by using a
$1650$ NVIDIA GeForce GPU, the minimum speed of this tracker is $40$
FPS for images with size $(448 \times 448)$, while it rushes to
$120$ FPS for $(56 \times 56)$ image size. Although a single object
tracking application is studied here, the proposed tracker may be
used for multiple-object tracking tasks. It is possible to track
several objects by adding an adequate number of trackers since the
suggested tracker has been implemented as an object in the code.
\textcolor{black}{The speed of the proposed tracker is given in Table.
\ref{table_speed} for further analysis. This table illustrates the
maximum speeds of the employed trackers obtained on the same
hardware. As it is seen in this table, the maximum speed of the
proposed tracker is $120$ FPS, which is clearly much better than
that of conventional trackers. Furthermore, comparing to the two
trackers with higher running speed than that of the proposed
tracker, the superiority in performance justifies the use of the
proposed tracker which is definitely implementable in real time
applications.}
\begin{table}
    \centering
    \caption{\textcolor{black}{This table reports the speeds of the trackers each obtained by running the corresponding tracker several times in a simple tracking task and saving the best running time. Note, in the mentioned task a small object is considered to be tracked to further help some of the trackers to be faster.}}
    \begin{tabular}{cccccccccc}\hline
        \tiny{Tracker} & \tiny{Proposed} & \tiny{BOOSTING} & \tiny{MIL} & \tiny{KCF} & \tiny{TLD} & \tiny{MF} &  \tiny{GOTURN} & \tiny{MOSSE} & \tiny{CSRT}\\\hline
        \tiny{Max Speed (fps)}& \scriptsize{120}    & \scriptsize{25} & \scriptsize{12} & \scriptsize{35}   & \scriptsize{10}   & \scriptsize{87}   & \scriptsize{100}  & \scriptsize{800}  & \scriptsize{23} \\\hline
    \end{tabular}
    \label{table_speed}
\end{table}

\subsection{The Failure Analysis}
Considering the high importance of failure rates among trackers, in this subsection, a monte-carlo simulation has been studied to further investigate the proposed tracker performance. Since almost all of the trackers are highly sensitive to the initialization, each scenario is repeated for $10$ times taking different initial conditions for the trackers. Not only does this take various initializations into account but also it may challenge the probable randomness in the trackers. In the beginning for the first time, the initialization is performed manually. Then, for the rest of the $9$ simulations, this initial condition is changed stochastically while considering a limited bound of randomness. Moreover, each scenario has been divided into three or four parts each of which is associated with a specific challenging task such as a fast movement, an occlusion, or a big change in the object size. Note that except for the last scenario, the results in the previous section are given for a part of a video. The whole videos, however, have been taken in this subsection to further consider various challenges such as camera movement.

\begin{table}
    \centering
    \caption{Results of the monte-carlo simulation given to further analyze the proposed tracker in different situations while taking failure rates into consideration}
    \begin{tabular}{ccccccccc}\hline
        \scriptsize{Proposed} & \scriptsize{BOOSTING} & \scriptsize{MIL} & \scriptsize{KCF} & \scriptsize{TLD} & \scriptsize{MF} &  \scriptsize{GOTURN} & \scriptsize{MOSSE} & \scriptsize{CSRT}\\\hline
        \footnotesize{(0,0,1)}  &\footnotesize{(0,2,1)} &\footnotesize{(0,1,1)} &\footnotesize{(0,4,3)} &\footnotesize{(0,4,3)} &\footnotesize{(0,2,0)} &\footnotesize{(0,3,3)} &\footnotesize{(0,4,3)} &\footnotesize{(2,4,3)} \\\hline
        \footnotesize{(0,1,1)}  &\footnotesize{(0,2,0)} &\footnotesize{(1,2,2)} &\footnotesize{(0,4,3)} &\footnotesize{(0,3,3)} &\footnotesize{(0,2,1)} &\footnotesize{(0,1,1)} &\footnotesize{(0,4,3)} &\footnotesize{(4,3,3)} \\\hline
        \footnotesize{(0,2,1)}  &\footnotesize{(0,3,1)} &\footnotesize{(0,2,2)} &\footnotesize{(0,3,3)} &\footnotesize{(0,4,3)} &\footnotesize{(0,2,1)} &\footnotesize{(0,1,3)} &\footnotesize{(0,3,3)} &\footnotesize{(1,3,3)} \\\hline
        \footnotesize{(0,1,3)}  &\footnotesize{(0,2,1)} &\footnotesize{(0,3,3)} &\footnotesize{(0,4,3)} &\footnotesize{(1,4,3)} &\footnotesize{(0,2,1)} &\footnotesize{(0,1,2)} &\footnotesize{(0,4,3)} &\footnotesize{(4,4,3)} \\\hline
        \footnotesize{(2,0,3)}  &\footnotesize{(0,1,3)} &\footnotesize{(0,2,3)} &\footnotesize{(0,4,3)} &\footnotesize{(1,3,3)} &\footnotesize{(0,2,1)} &\footnotesize{(0,2,2)} &\footnotesize{(0,3,3)} &\footnotesize{(1,4,3)} \\\hline
        \footnotesize{(0,2,1)}  &\footnotesize{(0,1,3)} &\footnotesize{(0,1,3)} &\footnotesize{(0,4,3)} &\footnotesize{(1,4,3)} &\footnotesize{(0,2,3)} &\footnotesize{(0,1,3)} &\footnotesize{(0,4,3)} &\footnotesize{(2,4,3)} \\\hline
        \footnotesize{(1,0,3)}  &\footnotesize{(0,2,1)} &\footnotesize{(0,2,3)} &\footnotesize{(0,4,3)} &\footnotesize{(0,4,3)} &\footnotesize{(0,2,0)} &\footnotesize{(0,4,3)} &\footnotesize{(0,3,3)} &\footnotesize{(4,4,3)} \\\hline
        \footnotesize{(0,1,3)}  &\footnotesize{(0,2,3)} &\footnotesize{(0,1,3)} &\footnotesize{(0,4,3)} &\footnotesize{(0,2,3)} &\footnotesize{(0,2,3)} &\footnotesize{(0,2,3)} &\footnotesize{(0,4,3)} &\footnotesize{(2,4,3)} \\\hline
        \footnotesize{(0,1,1)}  &\footnotesize{(0,1,1)} &\footnotesize{(0,2,1)} &\footnotesize{(0,3,3)} &\footnotesize{(1,2,3)} &\footnotesize{(0,2,1)} &\footnotesize{(0,1,2)} &\footnotesize{(0,3,3)} &\footnotesize{(3,4,3)} \\\hline
        \footnotesize{(0,0,1)}  &\footnotesize{(0,1,2)} &\footnotesize{(0,2,2)} &\footnotesize{(0,4,3)} &\footnotesize{(0,4,3)} &\footnotesize{(0,2,1)} &\footnotesize{(0,2,1)} &\footnotesize{(0,4,3)} &\footnotesize{(4,4,3)} \\\hline
    \end{tabular}
    \label{tableexample}
\end{table}

To illustrate the results, Table \ref{tableexample} is presented in which each raw shows the results of a simulation. Furthermore, the three numbers in each element present the corresponding scenario result in that simulation (e.g. in $(0,2,1)$, $2$ stands for the number of failures in the second scenario in the corresponding simulation). Before getting into the detail of the results, let's investigate the challenges in each scenario first. The first scenario has been divided into four parts including various object views, changing in the object size, shaking camera, and cluttered background. The second scenario is associated with various object sizes while having different object views, moving camera, fast movement while becoming small quickly, and a short-term full occlusion. Finally, the third scenario challenges are the moments in which the disk is hit by the player. In contrast to the other two scenarios, three failure points are considered in this scenario since there are three hit points. Note that, when a tracker fails at each of the failure points, it will fail for the rest of the points, as well. As a result, having four failures for a tracker means that it has been a failure at the beginning of the tracking process. Taking the results given in Table \ref{tableexample}, the designed tracker has an acceptable performance through the challenges mentioned in the first scenario. An important point here is that despite the cluttered background, the proposed tracker has performed well when dealing with similar environmental structures like billboards in the background. Besides, in the second scenario, the proposed tracker has performed the tracking successfully with no failures for $\%40$ of the simulations while the others including the CNN based GOTURN tracker have failed at least one time in each simulation. This means that considering the challenges mentioned in the second scenario, one common failure among all the trackers except the proposed one is the short-term occlusion. Thus, as it has been claimed, the proposed tracker may even handle short-term occlusions. Finally, for the third scenario, the proposed tracker stands with $18$ failures while the GOTURN as the other CNN based tracker has at least $23$ failures indicating that the designed tracker has significant advantages compared to the GOTURN tracker. In conclusion, the promising results of the designed tracker demonstrate some of its superiorities over the commonly employed trackers in the literature considering its acceptable performance through various challenging situations.

\begin{figure}[]
    \centering
    \includegraphics[width=4.5in]{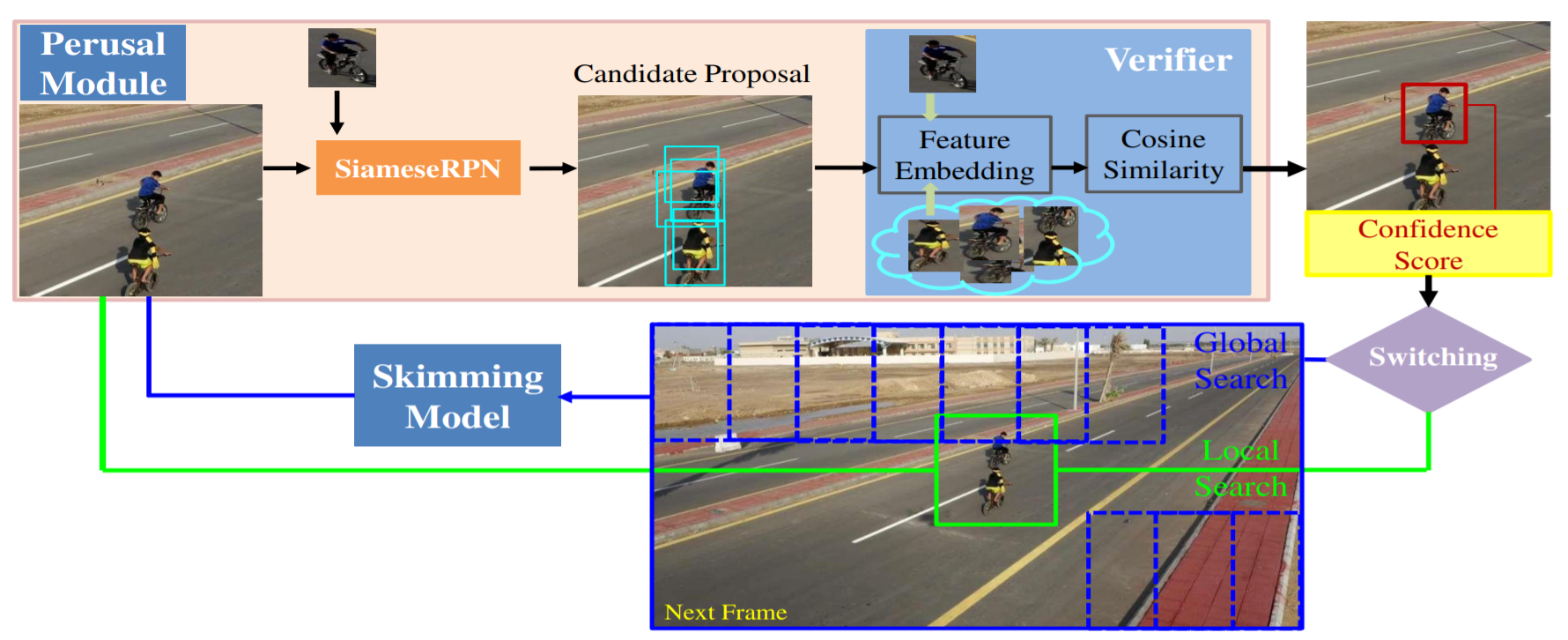}
    \caption{Skimming-Perusal’ long-term tracking framework presented by \cite{SPLT}}
    \label{fig_splt}
\end{figure}

\begin{figure}[]
    \centering
    \subfloat[]{\includegraphics[width=2.3in]{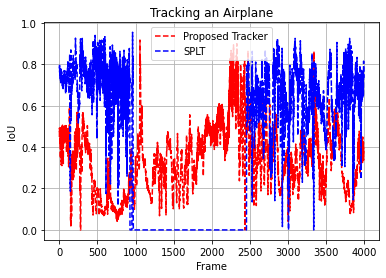}
        \label{fig2_1}}
    \subfloat[]{\includegraphics[width=2.3in]{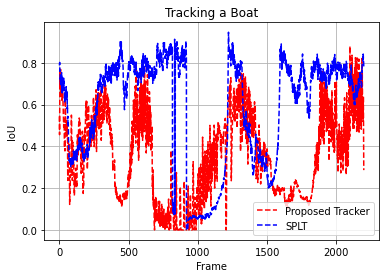}
        \label{fig2_2}}\\
    \caption{This figure illustrates the results of the proposed tracker and the SPLT on two long-term tracking scenarios.}
    \label{fig_res1}
\end{figure}

\textcolor{black}{
\subsection{Performance of the Proposed Tracker on the LaSOT Dataset}
In this part, as the final study of the proposed tracker
performance, another dataset called LaSOT\cite{lasot} is examined to
train the proposed architecture on it. Furthermore, the tracker is
compared to the state-of-the-art and recently proposed tracker
called "SPLT"~\cite{SPLT} to get better insight on the results.
Although a complete comparison has been made on the commonly used
trackers, performing this comparison may help to further demonstrate
the proposed tracker capabilities. Furthermore, to avoid worries
related to implementing this tracker on another dataset which may
downgrade its performance, both trackers are trained on the LaSOT
dataset where SPLT has been trained and tested on it.
}
\textcolor{black}{
The overview of SPLT tracker~\cite{SPLT} is shown in Fig.
\ref{fig_splt}, where the authors have claimed a robust and realtime
performance on this structure. The SPLT tracker is based on the
proposed skimming and perusal modules. Furthermore, a verifier based
on ResNet50 is utilized to infer whether the object is absent or
present. This outcome is employed to choose between performing a
global search or a local one. The results given in~\cite{SPLT} are
very promising, indicating the SPLT tracker could be named as the
state-of-the-art tracker in this field. Although by using strong
hardware it may perform realtime, due to its heavy architecture, it
is not comparable with our proposed tracker in terms of speed. Our
running observation shows that this tracker is at least five time
slower than that of our proposed tracker.}
\textcolor{black}{
Consequently, the comparison may challenge the proposed tracker
accuracy, and not its speed. In this regard, considering the
proposed tracker light structure, it is not expected to get higher
accuracy. However, an acceptable performance compared to the SPLT
may demonstrate its capabilities in both long-term tracking and
other challenging situations. For this means, four scenarios are
taken into account each having different challenges. These scenarios
correspond to "airplane-9", "boat-4", "bottle-18", and "cat-1" test
videos in LaSOT.}
\textcolor{black}{
The first two scenarios are considered as long-term tracking with
various challenges such as short-term occlusion, diverse point of
view, frequent change in the object size, different movement speed,
and zooming in or out. The results of using the proposed tracker and
the SPLT are shown in Fig.~\ref{fig_res1}. In the first figure on
the left side, an improper initialization is considered for the
proposed tracker to further make it more challenging to continue on
the tracking. Since the SPLT tracker could not handle the occlusion,
it is tried to make the situation even harder for the proposed
tracker to see if it can handle such situation. As it is seen in
this figure, even though the IoU corresponding to the proposed
tracker has been decreased to nearly a failure, it could properly
handle the occlusion and enhance the IoU gradually, as it can be
seen in Fig.~\ref{fig_res1}~(a) from frame number $1000$ to $2500$.
Here, an important note to ponder is that although the proposed
tracker is much faster and it does not have a global searching
module, it could properly handle a short-term occlusion, and it
could maintain a constant IoU average over the whole tracking task.
Fig.~\ref{fig_res1}~(b)  shows the results corresponding to a task
of tracking a boat. In this task, challenges are similar to the
previous one except that the camera has much higher movement
compared to that of the previous task. Compared to the SPLT, the
proposed tracker performs properly and in some situations even
better than that of the SPLT. The outstanding characteristic of the
proposed tracker is that it can keep low IoU to a small amount
without failure, and more importantly, it consistently improves the
predictions without having a global search module. }

\begin{figure}[]
    \centering
    \subfloat[]{\includegraphics[width=2.3in]{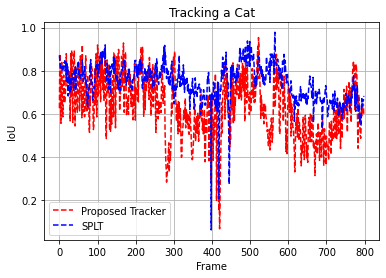}
        \label{fig3_1}}
    \subfloat[]{\includegraphics[width=2.3in]{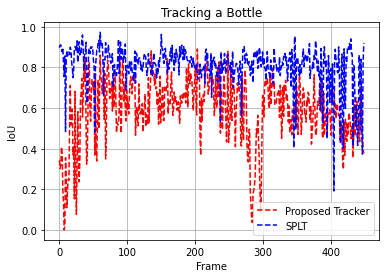}
        \label{fig3_2}}\\
    \caption{The first result corresponds to a video of a cat in which the cat sometimes changes its position suddenly while the camera zooms in or out. The other outcome reports the results of the two tracker on tracking an object while having severe fluctuations in the camera caused by its movement. This results in a completely blurred images and abrupt change in the object position.}
    \label{fig_res2}
\end{figure}
\textcolor{black}{
The other two scenarios, on the other hand, contains blurry images
caused by the camera rapid movements, abrupt change in the object
position, and zooming in and out. Results are reported in Fig.
\ref{fig_res2}. As it is seen in this figure, dealing with the
challenges in these scenarios, the proposed tracker performed even
better than the former two scenarios. Note that the improper
initialization is considered deliberately as these cases as well, to
examine the tracking task failure or downgrading in the performance.
The aim of this type of initialization is to further demonstrate the
tracker's capability in enhancing the results. As shown in Fig.
\ref{fig3_2} by considering the results obtained in these two
scenarios, the performance of the proposed tracker is comparable and
even better than that of the SPLT results in terms of accuracy.
Finally, our claim about handling short-term occlusions is verified
by both the failure study and this comparison. }

\section{Conclusions}
In this paper, a training approach is presented to precisely
localize the object in an image. The ROI and the center point
matrices are used to determine the region of interest and the most
probable center point of the object in the image, respectively.
Furthermore, considering various input sizes, a single
multiple-model CNN is proposed. This model contains a flexible
structure with distinct and common layers to keep the performance
high regardless of the object sizes. \textcolor{black}{To demonstrate
the capability of both the training method and the suggested model,
a tracking application has been considered, and two comparative
studies over eight OpenCV trackers including the GOTURN, and a
recently proposed tracker called SPLT are performed. These studies
reveal the applicability of the suggested tracker and the proposed
training approach.} Note that there are some limitations to the
suggested method. For instance, the proposed approach does not
distinguish between objects, and it has a part to extract distinct
features. Moreover, even though the issues like short-term occlusion
may be well handled compared to the other trackers, it is
dispensable to have a specific module to handle long-term occlusions
where features may be lost. Consequently, our current reserach is
focused on adding a classifier module in addition to a recurrent
neural network module to address the above mentioned challenges.

\section*{References}

\bibliographystyle{ieeetr}
\bibliography{Refarticle}
\end{document}